\documentclass[runningheads]{llncs}
\usepackage{graphicx}
\usepackage{xcolor}
\usepackage{hyperref}

\usepackage{booktabs}
\usepackage{subcaption}
\usepackage{amsfonts}
\usepackage{diagbox}
\usepackage[T1]{fontenc}
\begin{document}

 \author{Akansha Kalra, Soumil Datta, Ethan Gilmore, Duc La,\\ Guanhong Tao, Daniel S. Brown}
\authorrunning{A. Kalra et al.}

\institute{University of Utah \\ \email{\{akansha.kalra, soumil.datta, guanhong.tao, daniel.s.brown\}@utah.edu}
}

\title{Dataset Poisoning Attacks on Behavioral Cloning Policies}
%
%
\maketitle              
\begin{abstract}
Behavior Cloning (BC) is a popular framework for training sequential decision policies from expert demonstrations via supervised learning. As these policies are increasingly being deployed in the real world, their robustness and potential vulnerabilities are an important concern. In this work, we perform the first analysis of the efficacy of clean-label backdoor attacks on BC policies. Our backdoor attacks poison a dataset of demonstrations by injecting a visual trigger to create a spurious correlation that can be exploited at test time. We evaluate how policy vulnerability scales with the fraction of poisoned data, the strength of the trigger, and the trigger type. We also introduce a novel entropy-based test-time trigger attack that substantially degrades policy performance by identifying critical states where test-time triggering of the backdoor is expected to be most effective at degrading performance. We empirically demonstrate that BC policies trained on even minimally poisoned datasets exhibit deceptively high, near-baseline task performance despite being highly vulnerable to backdoor trigger attacks during deployment. Our results underscore the urgent need for more research into the robustness of BC policies, particularly as large-scale datasets are increasingly used to train policies for real-world cyber-physical systems. Videos and code are available at \url{https://sites.google.com/view/dataset-poisoning-in-bc}.

\keywords Imitation Learning \and Dataset Poisoning \and Backdoor Attacks

\end{abstract}
\section{Introduction}
\label{sec:intro}
When learning sequential decision-making tasks, there is often a trade-off between using Reinforcement Learning (RL) or using Imitation Learning (IL). 
While RL provides a way to train a policy without requiring any more supervision than a reward signal, it introduces the problem of balancing exploration and exploitation and is often quite sample inefficient, requiring large amounts of interactions with the environment.
The main principle of RL is an iterative process of trying various actions to discover what works~\cite{sutton1998reinforcement}. However, in real-world cyber-physical systems, this trial-and-error approach can result in dangerous unsafe situations and large computational costs.
Additionally, RL requires access to a carefully designed or learned reward function which is often hard to specify correctly for complex tasks~\cite{amodei2016concrete,hadfield2017inverse}, leading to potentially unwanted behaviors~\cite{tien2023causal}.

By contrast, IL seeks to learn policies from expert demonstrations. The most common and popular method for IL is behavioral cloning (BC), where a policy is directly learned from state-action pairs \cite{ross2011reductionimitationlearningstructured,torabi2018behavioralcloningobservation}.
This approach has recently gained much success in real-world deployments and has been shown to scale well to large datasets and modern deep learning architectures \cite{chi2023diffusionpolicy,florence2021implicit,zhao2023learningfinegrainedbimanualmanipulation}.
Learning from demonstrations via supervised learning allows IL methods to learn policies without requiring interaction with the environment, improving sample efficiency and bypassing the need for potentially dangerous exploration. 
However, it is known that these types of models are often susceptible to compounding errors when they encounter out of distribution states~\cite{ross2011reductionimitationlearningstructured}. This inherent vulnerability could be exploited by a malicious adversary through a well-timed attack. 
All the attacker may need to do is perturb the input observation to the policy by a tiny amount at the right moment to drive the policy out of distribution. Then compounding errors take effect, causing the policy to fail or perform poorly.

In this work, we study the vulnerability of BC to dataset poisoning attacks. Dataset poisoning attacks involve altering a training dataset to introduce a vulnerability into downstream models that are trained on this poisoned dataset~\cite{li2022backdoorlearningsurvey}. These attacks require only minimal edit access to a training dataset---something that is becoming increasingly possible as large datasets become publicly available~\cite{o2024open,khazatsky2024droid}.  We focus on clean-label backdoor attacks \cite{turner2018clean,turner2019labelconsistentbackdoorattacks}, a subclass of poisoning attacks where a small trigger, such as a 3x3 red patch, is embedded into training observations corresponding to a specific action (e.g., ``gas") without changing the actual labels (demonstrator actions). The goal is to create a ``backdoor effect": a scenario where a malicious actor introduces specific data points into a training dataset with the goal of causing a machine learning model to behave in a predetermined, undesirable way when presented with a specific ``trigger" during testing or inference, while otherwise behaving normally. Essentially, the attacker creates a hidden vulnerability (the backdoor) that can be activated by the trigger, allowing the attacker to manipulate the model's output for their benefit. These types of attacks work because the poisoned training data creates a spurious correlation between the visual trigger and the target action that is likely to be learned by any downstream model. 

To the best of our knowledge, we are the first to study the effectiveness of clean-label dataset poisoning attacks on behavioral cloning by exploring and analyzing the impact of different amounts of poisoned data, trigger strengths, and types of backdoor triggers. 
Our results show that BC policies are highly vulnerable to dataset poisoning attacks. Even a poisoning around 2.3\% of a training dataset is enough to create a strong backdoor effect. Given an effective backdoor, we then study the vulnerability of behavioral cloning policies to backdoor triggers during test time. Importantly, in sequential decision-making, not all states are equally critical. In car driving, for example, triggering a backdoor to trigger an acceleration action on a straight road segment may be benign, but triggering it during a sharp turn when the agent should turn left can push the policy out of distribution and amplify errors through covariate shift~\cite{ross2011reductionimitationlearningstructured}. 
This motivates our novel test-time trigger attack that uses a policy entropy threshold to trigger attacks only at critical states, making the best use of a limited attack budget. 







Our main contributions are the following:
(1) We present the first study and discussion of dataset poisoning in imitation learning;
(2) We perform a sensitivity analysis over trigger types, amount of poisoned data, and patch size for clean-label attacks on behavioral cloning policies;
(3) We demonstrate that our entropy-based test time attack significantly degrades performance at test time when compared with nominal baseline performance and randomized attacks. 


\section{Background \& Related Work}
\label{sec:related-work}


Imitation Learning (IL) has emerged as a powerful tool for training sequential decision-making policies by mimicking expert demonstrations, offering an alternative to reinforcement learning (RL) in scenarios where explicit reward functions are hard to specify or costly to obtain~\cite{ng1999policy,krakovna2020specification}. 
Among IL approaches, Behavioral Cloning (BC) remains the most widely used due to its simplicity and scalability. 
It frames policy learning as a supervised learning problem over expert state-action pairs \cite{ross2011reductionimitationlearningstructured,torabi2018behavioralcloningobservation}. 
BC has shown great success across cyber-physical systems such as autonomous driving \cite{bojarski2016endendlearningselfdriving,bc2018_autonomous}, and robotics manipulation \cite{chi2023diffusionpolicy,florence2021implicit}, often requiring significantly fewer interactions with the environment compared to traditional RL methods.

Early work on adversarial attacks on deep reinforcement learning identified that not all time steps are equally vulnerable to perturbation and that adversaries can achieve high effectiveness by injecting noise only at critical moments when the agent expects high reward~\cite{lin2017tactics,qiaoben2021strategically,russo2019optimalattacksreinforcementlearning,Li2023statebased}. These approaches demonstrate the ability, in the RL setting, to reduce attack frequency while maximizing attack impact, highlighting the importance of timing in adversarial attack strategies. 
However, these attacks require white-box access to the policy, access to a reward function, and access to a simulator to devise the attack strategy and are not applicable in learning from demonstrations settings, such as the ones we consider in this paper, where we assume black-box access to the policy and no access to a ground-truth reward function nor simulator when triggering attacks.

Despite its widespread adoption, relatively little is known about the robustness and security of IL policies. Prior work has examined adversarial perturbations in driving simulations that are generated on the fly using white-box access to the policy~\cite{hall2020studying} or used white-box policy access to design physical adversarial examples like black lines painted on the roadway~\cite{boloor2019simple}. Other work explores white-box adversarial attacks on modern diffusion policies~\cite{Chen2024DiffusionPA} as well as transfer attacks between models~\cite{patil2025vulnerablepolicyadversarialattacks}. However, all these approaches also assume white-box access to pre-trained BC policies and involve complicated optimizations to solve for adversarial pixel perturbations at test time. By contrast, we consider backdoor-triggered dataset poisoning within the context of BC and demonstrate the effectiveness of cheap, black-box attacks. 



Data poisoning and backdoor attacks have been extensively studied in the context of supervised learning, where adversaries inject malicious examples into the training data to degrade performance or implant a targeted behavior at test time~\cite{li2022backdoorlearningsurvey,gu2019badnets,turner2019labelconsistentbackdoorattacks,chen2017targetedbackdoorattacksdeep,blindbackdoors2021,hiddentriggerbackdoor_2020,turner2018clean}. These attacks are particularly dangerous because they do not require white-box access to model parameters when performing an attack. While there has been some recent work on backdoor attacks for reinforcement learning~\cite{kiourti2020trojdrl,chen2023bird,gong2024baffle,guo2025pnact}, to the best of our knowledge, we are the first to study dataset poisoning attacks on policies that are learned from datasets of demonstrations without access to reward functions. Our attacks only require minimal and simple edits to a dataset to inject a backdoor to compromise a dataset such that downstream policies trained on the poisoned dataset are vulnerable. Our test-time attacks are also computationally simple to apply (we simply apply a predetermined, deterministic trigger to the input) and highly effective despite requiring only black-box access to the outputs of the BC policy.



\section{Methodology}
\label{sec:method}
We consider the standard Learning from Demonstration (LfD) setting where an attacker has access to a set of expert demonstrations $D= \{ (s,a), \ldots \}$ of state-action pairs $(s,a)$ that will be used by a downstream imitation learning algorithm to learn a policy~\cite{ross2011reductionimitationlearningstructured,torabi2018behavioralcloningobservation}. 
The objective of the attacker is to poison the training data by injecting a trigger into the training dataset, so that, at test time, it will enable the attacker to control a BC policy trained on the poisoned dataset by triggering a chosen target action, $a_{target}$.

\subsection{Threat Model}
Before describing our attacks, we first clearly specify our threat model.

\vspace{2mm} \noindent \textbf{Adversary's Goal}: By injecting a clean-label backdoor into the training data, the attacker aims to induce failures in the learned BC policy when the backdoor is triggered at test time. For example, the attacker could use the trigger to cause an autonomous car to take an acceleration action rather than a brake action at a red light, potentially leading to dangerous or catastrophic behavior.


\vspace{2mm} \noindent \textbf{Adversary's Knowledge and Capabilities}: The attacker has complete access to the dataset for poisoning, but has no direct control over nor privileged white-box access to downstream policy models that are trained on the poisoned  training data. At test time, the attacker has only black-box access to the policy model outputs and can apply only a limited budget of test-time attacks by injecting the backdoor trigger into the visual observation of the pre-trained policy. 
We note that these perturbations are limited to the visual observation space (the adversary has no direct action manipulation ability).

\subsection{Dataset Poisoning}
\label{dataset_poisoning}

In this section we describe our clean label poisoning-based backdoor attack.
The attack operates on an existing dataset $D= \{ (s,a), \ldots \}$ of state-action pairs $(s,a)$, where $s$ is the state observation and $a$ is the demonstrator's action. Given a desired backdoor target action $a_{\rm target}$, the attacker edits $D$ to create a poisoned dataset $D_p$. The poisoned dataset $D_p$ is formed by taking some number of state-action pairs $(s,a)$ from $D$ with $a=a_{\rm target}$ and then adding a backdoor trigger to $s$ to create a poisoned state action pair $(\tilde{s},a)$.
The goal of the attacker at test time is to minimize $J(\pi_{\rm bc}) = \mathbb{E}_{\pi_{\rm bc}}\left[\sum_{t=0}^T r_t \right]$, where $\pi_{\rm bc}$ is the stochastic policy trained on $D_p$.
In practice, we poison the dataset by embedding backdoor triggers formed by n$\times$n patches of pixels into the image observations, $s$. Note that because this is a clean-label attack, we perturb only the state observation, $s$, without ever altering the expert actions, $a$, when creating $D_p$.
In our experiments, we analyze the effect of clean-label poisoning on BC policies by varying the fraction of data poisoned, the strength of the trigger via patch size, and the trigger type.

\subsection{Test-Time Trigger Attacks}
In addition to dataset poisoning, we study the effect of the timing of backdoor trigger attacks during test-time execution, given an attack budget $B$ of test-time backdoor injections. Our insight is that certain states are better to attack than others. For instance, in case of car-driving, applying the backdoor trigger on a straight road segment observation to trigger the gas action may be inconsequential, whereas doing so during a sharp curve in road---such as when a left turn is needed---can lead to out-of-distribution states and lead to compounding errors and catastrophic failure. We compare two strategies for determining when to inject the backdoor trigger at test time: 

\vspace{2mm} \noindent \textbf{Random Trigger Timing:} This baseline simply selects $B$ timesteps uniformly at random to trigger the backdoor attack.

\vspace{2mm} \noindent \textbf{Entropy-Based Trigger Timing:}
We also propose and study a more sophisticated way of deciding when to trigger the backdoor attack. Our idea is to use the entropy of the action output distribution of the BC policy to identify critical states where we expect the attack to be more effective. 
Mathematically, given a trained stochastic BC policy $\pi_{\rm bc}(a\mid s)$, trained on the  poisoned dataset $D_p$, we assume the attacker has black-box access to the model outputs---i.e., they can query the policy to obtain action probabilities, but cannot inspect or modify internal model parameters. At test time, for each input state $s$, the attacker computes the BC policy action entropy: $\mathcal{H}(\pi(\cdot \mid s)) = - \sum_{a \in \mathcal{A}} \pi_{\rm bc}(a \mid s) \log \pi_{\rm bc}(a \mid s)$ where $\mathcal{A}$ refers to the action space.\footnote{We only consider discrete action spaces in this paper, but our proposed entropy-based attack can easily be extended to continuous action distributions (e.g., Gaussian policies) by computing or estimating the differential entropy.}
Observations with entropy below a specified threshold where the output of the policy $\pi(a \mid s) \neq a_{target}$ are selected as for inserting adversarial triggers until the budget $B$ is exhausted. The entropy threshold is attacker-defined and can be tuned at test time. The intuition behind this trigger timing strategy is that states with low entropy imply that there are only a small number of appropriate actions, and the smaller the entropy, the more likely it is that only one action is appropriate. If this action is not equal to the target action $a_{target}$, then triggering the policy to output $a_{target}$ is likely to cause the wrong state transition and degrade policy performance by sending the policy out of the distribution of demonstrations in the training data. 






\section{Experiments and Results}
We begin by first outlining our experimental setup and then turn to the investigation of the key experimental questions in the following subsections.

Throughout our experimental framework, we use the Car Racing environment from OpenAI gym~\cite{brockman2016openaigym} comprising of high-dimensional visual observations in the form of 96×96-pixel RGB images and a discrete action space comprised of 5 actions: do nothing, gas, turn left, turn right, and brake.
Its high-dimensional visual inputs permit the introduction of spatially constrained triggers, while its discrete action space allows for precise measurement of behavioral deviations and reward variation under distribution shift.
Following prior works~\cite{brown2019betterthandemonstratorimitationlearningautomaticallyranked,chen2021learning} that use synthetic demonstrations to model an expert, we utilize a PPO agent \cite{schulman2017proximalpolicyoptimizationalgorithms}, trained for 1M steps to generate the clean demonstration datasets used in our experiments.

As discussed in Section~\ref{dataset_poisoning}, we embed backdoor triggers into the clean dataset to create the poisoned dataset. We then train Behavior Cloning (BC) policies on the poisoned datasets using a classical convolutional neural network architecture. 
For training the BC policies, we use Adam optimizer with learning rate equal to 0.001 and we leverage the classical feedforward CNN architecture neural network with 3 convolutional layers each with 3x3 kernels followed by ReLU and pooling layers, which are then followed by 3 fully connected layers with ReLU as activation except at the last layer where we use softmax to get the final probability distribution across all possible actions. 

\subsection{How vulnerable are BC policies to clean-label backdoor attacks under different amounts of dataset poisoning?}

To answer this question, we perform clean-label poisoning by modifying observations associated with the gas action in the expert dataset, embedding a solid 3×3 red patch in the top-left corner of the image. 
This setup allows us to assess both the effectiveness and efficiency of the backdoor trigger.
The fraction of state observations (corresponding to gas-actions) that are poisoned is incrementally varied between 0\% and 100\%. 
For each specified poisoning percentage, we construct the corresponding poisoned dataset and train a separate BC policy. We assess the learned policy’s performance in two ways: (i) by measuring the mean episode reward during standard test-time rollouts, and (ii) the backdoor control rate, defined as the proportion of times the presence of the trigger induces the gas action.

\begin{figure}[t]
\centering
\includegraphics[width=\textwidth]{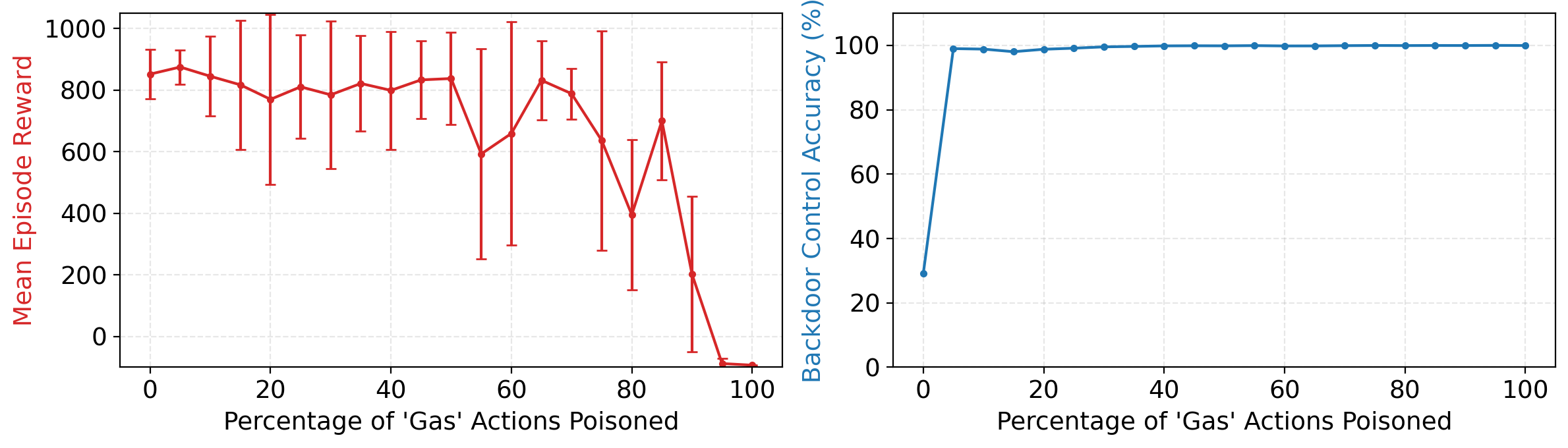}
\caption{\textbf{Sensitivity to the percentage of poisoned training data.} The figure displays mean episode reward (left) and backdoor control rate (right) for varying poisoning percentages, based on a fixed 3×3 red patch on gas-labeled frames, averaged across 5 seeds with 10 rollouts per seed with error bars denoting standard deviation. We show that BC policies can achieve near-baseline performance despite being poisoned while even the minimal amount of poison injected lends complete controllability to the attacker.} \label{fig:ethan-red-patch}  
\end{figure}

Our empirical results in Fig. \ref{fig:ethan-red-patch} show that the backdoor is highly effective: with just 5\% poisoning of 'gas' actions, which corresponds to only 2.31\% of the overall dataset, the backdoor control rate increases from chance to approximately 100\%, and remains at this level thereafter. This indicates that the red patch functions as a highly salient and learnable trigger. 
Meanwhile, the mean reward remains largely stable until the poisoning ratio surpasses $70$–$80$\%, after which the reward drops sharply, culminating in full collapse at 100\% poisoning. This suggests that extreme poisoning levels introduce training instability, likely due to the dominance of poisoned correlations in the data distribution. 
This indicates that BC policies are highly susceptible to clean-label backdoor attacks, even at minimal poisoning levels. A small fraction of poisoned training data is sufficient to grant the adversary near-perfect control at test time, while overall task performance at test time rollouts remains deceptively high until poisoning exceeds a critical threshold. 

Our results reveal a critical vulnerability: near-baseline task performance persists even when the BC policy is fully compromised by a backdoor trigger. This means that traditional evaluation metrics are insufficient—the model may appear robust while being entirely controllable under adversarial conditions. 

\subsection{Which visual properties of synthetic triggers most influence the backdoor trigger efficacy?}

\begin{figure}[t]
    \centering
    \begin{subfigure}[t]{0.24\textwidth}
        \centering
        \includegraphics[width=\linewidth]{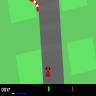}
        \caption{\scriptsize Clean}
    \end{subfigure}%
    \hfill
    \begin{subfigure}[t]{0.24\textwidth}
        \centering
        \includegraphics[width=\linewidth]{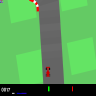}
        \caption{\scriptsize Gaussian}
    \end{subfigure}%
    \hfill
    \begin{subfigure}[t]{0.24\textwidth}
        \centering
        \includegraphics[width=\linewidth]{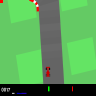}
        \caption{\scriptsize Red}
    \end{subfigure}
    \caption{\textbf{Visual comparison of clean and poisoned frames using different attack types.}  All attacks shown are a 3$\times$3 patch in the top left corner of the visual observation.}
    \label{fig:poisoned-comparison2}
\end{figure}

To understand the influence of visual characteristics on backdoor trigger efficacy, we compare two types of synthetic patch triggers: a fixed-color square, a Gaussian noise patch. These trigger variants are visualized within poisoned observations in Fig.~\ref{fig:poisoned-comparison2}.

The fixed-color square consists of a spatially fixed red patch with high visual contrast, making it easy to learn but also more perceptible. 
The Gaussian noise patch is sampled from a normal distribution $\mathcal{N}(127, 30^2)$, clipped to $[0, 255]$. We generate it once and reuse the same instance on every poisoned frame to provide a consistent correlation (resampling for each frame would make the trigger stochastic and much less likely to induce a backdoor effect). The main benefit of the Gaussian noise patch is a lower chance of detection by simple defenses, while its drawback is that it may be harder to learn at lower poisoning rates and may require a higher rate to be more effective.



Our results in Table ~\ref{table:patch-type} show that even with only 5\% of gas actions poisoned, both red and Gaussian patches achieve high backdoor control. The red patch yields the highest control and reward due to its strong contrast and consistency. The Gaussian patch is less effective at low poisoning rates, likely due to its lower visual salience. We also investigated applying a color-space shift patch~\cite{jiang2023color} for dataset poisoning, but did not find it to yield a stronger attack than Gaussian or Red patch so we focus on only Gaussian and red patch backdoors in the rest of the paper. Details and results for the color-space shift trigger are included in Appendix~\ref{app:more-patch-type}.

\begin{table}[t]
\vspace{-2mm}
\centering
\caption{\textbf{Comparison of the efficacy of backdoor patch types at 5\% poisoned gas actions averaged across 10 seeds.} Backdoor control accuracy is the percentage of poisoned observations predicted as ``gas''. The red patch performs visibly better in terms of backdoor control accuracy compared to the Gaussian noise patch, and remains comparable in terms of mean reward.}
\vspace{0.1cm}
\label{table:patch-type}
    \begin{tabular}{lcccccc}
    \toprule
    Patch Type &&& Mean Reward (± SE) &&& Backdoor Control Acc. (\%) (± SE)\\
    \midrule
    None          & &  & 841.28 ± 1.676 &&& 30.80 ± 0.003 \\
    Red Pixels     &  &     & 838.99 ± 1.278 &&& 96.45 ± 0.001 \\
    Gaussian Noise &&& 836.56 ± 1.769 &&& 80.35 ± 0.013 \\
    \bottomrule
    \end{tabular}
\end{table}

\subsection{How important is timing of backdoor triggers?}
We examine the impact of backdoor activation on a poisoned BC policy within the context of sequential decision-making by constraining the test-time attack budget to $B=100$ and evaluating two distinct backdoor triggers: a red patch and a Gaussian patch, applied to their corresponding backdoor-poisoned BC policies trained on respective datasets with 5\% of all gas actions poisoned.
We compare and contrast the random test time backdoor trigger attack with our novel entropy-based attack that injects an attack at test-time without needing any privilege information about the environment dynamics nor access to weights of the pre-trained BC model.
We solely utilize the outputs of the pre-trained backdoor poisoned model for our entropy-based attack. 


\begin{table}[t]
\centering
\caption{\textbf{Evaluation of test-time trigger attacks.} The table  depicts performance of BC policy across 100 rollouts (mean and standard error), under different backdoor triggers during test time given an attack budget of 100 and specified entropy threshold as 0.005.}
\vspace{0.1cm}
\begin{tabular}{|c|c|c|c|}
\hline
\diagbox{\textbf{Trigger Type}}{\textbf{Evaluation}} &
\textbf{Unattacked} & 
\textbf{Random} & 
\textbf{Entropy} \\ \hline
 Red Patch & 832.74 (12.84) & 794.96 (15.12) & -4731.69 (1178.56) \\ \hline
Gaussian Patch & 858.38 (9.31)  & 841.96 (8.30) & 577.43 (16.46) \\ \hline

\end{tabular}

\label{table:test-time-trigger-table}
\end{table}

Our results in Table \ref{table:test-time-trigger-table} demonstrate that our entropy-based attacks significantly degrade the performance of the BC backdoor poisoned policy as compared to when backdoor is not activated at test time.
We also show that our entropy-based attack outperforms the random attack baseline for test-time attacks by a highly significant margin of more than 100\% degradation in case of the red patch and a moderate  31\% degradation in case of the gaussian patch respectively. This justifies the need for intelligent backdoor trigger strategies at test time.

\subsection{How does  varying the size of the backdoor patch affect the trade-off between stealth and efficacy in dataset poisoning?}
\label{sec:patch-size-main}
Prior work \cite{gu2019badnets} highlights a trade-off between backdoor stealth and effectiveness: larger visual triggers can induce stronger spurious correlations but risk occluding task-relevant features, while smaller patches are less detectable but typically require higher poisoning rates.
To investigate this trade-off, we fix the poisoning rate at 5\% of gas actions poisoned, and vary the square ($N\times N$) patch size where $N\in[1,96]$ pixels. 
Each trigger is a red square placed in the top-left corner of the frame, targeting the "gas" action. For each patch size, we train  BC models for 5 separate seeds and report the average results over all seeds.


\begin{figure}
\centering
\includegraphics[width=\textwidth]{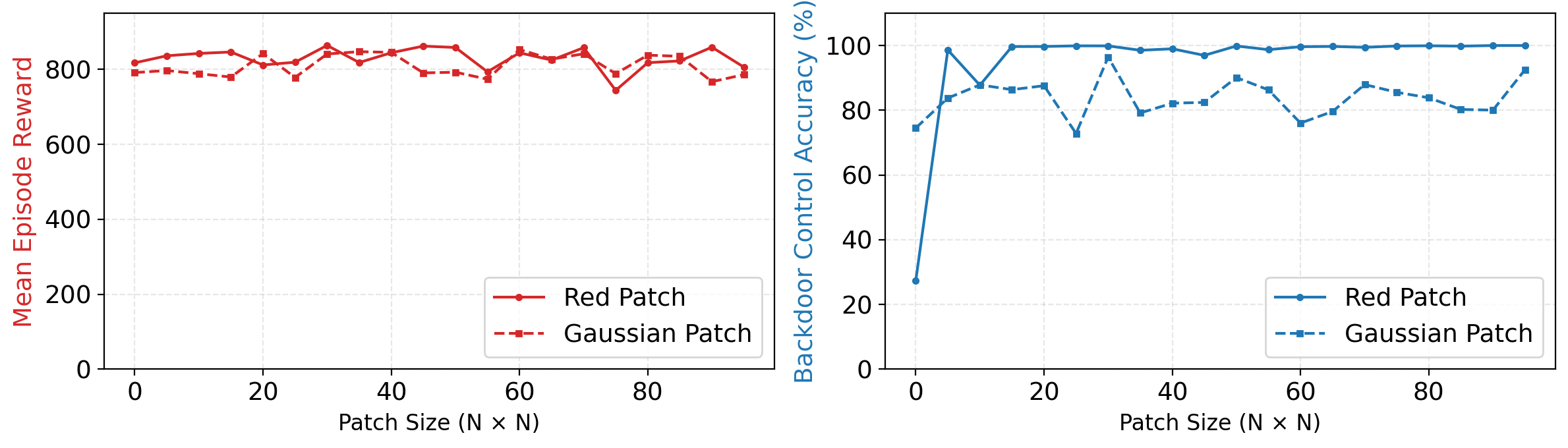}
\caption{\textbf{Patch size sensitivity for red pixel and Gaussian noise patches.} All models are trained with 5\% of gas-labeled frames poisoned using an $N{\times}N$ red square patch. We see that we have similar reward performance for both patch types across all $N$. We also see that the backdoor effect rises at a very small $N$ and stays fairly stable afterwards. The Gaussian patch performs worse in terms of control rate as a small poisoning rate is not enough for the policy to learn a more subtle trigger.} \label{fig:patch-size-attacks}
\end{figure}

Our results in Figure~\ref{fig:patch-size-attacks} indicate that small to moderately-sized red patches, e.g., sizes from 5x5 to 25x25 can slightly improve task reward, likely by reducing policy uncertainty in gas predictions. Gaussian patches show similar scaling with patch size, though with weaker and more variable control.
We find that trigger effectiveness plateaus around 5×5---larger patches offer no additional benefit and may increase detectability---highlighting the minimal patch size needed for effective backdoor activation.
We repeat the sweep at a higher poisoning rate of 20\% of gas-labeled frames to rule out a low-rate artifact and observe the same results. More details on this are included in the Appendix \ref{app:more-patch-size}.


\section{Discussion and Conclusion}
Our work reveals a fundamental vulnerability in BC policies: their susceptibility to very simple, low-budget clean-label dataset poisoning attacks. Our results show that 
BC policies compromised by backdoors may still achieve near-baseline performance unless the backdoor is actively triggered. Furthermore, our results show that even with as little as 2.31\% poisoned training data, BC policies can be manipulated to respond predictably to small visual triggers. 

Our experiments using a simple red patch as the backdoor trigger, in particular, demonstrates how minimal visual signals can yield near-complete control of model behavior to an adversary. This is a concerning result given the increasing deployment of IL models in real-world systems and the potential of using physical objects to trigger backdoors rather than requiring an adversary to hack into the visual feed of a cyber-physical system. This also highlights a major weakness of standard evaluation BC procedures, which can overlook a policy’s hidden vulnerability to adversarial manipulation.

Our results also provide interesting evidence that the efficacy of dataset poisoning attacks on BC policies may be more sensitive to  percentage of poisoned actions in the dataset rather than size of poison patch injected into the dataset.
We believe this might be due to the fact that frequent occurrence of the visual trigger across training dataset reinforces the association between the trigger and the target action, increasing the likelihood of successful backdoor activation at test time, irrespective of patch size. Future work should see if these trends hold across other environments and other imitation learning algorithms.

Our results illustrate a clear trade-off between the efficacy and applicability of different patch types used for backdoor attacks. We find that the red patch, with its high contrast is ideal for achieving strong backdoor control with minimal poisoning, but its overt nature severely limits its stealth in practical deployments.
Conversely, the Gaussian noise patch, designed to blend into the input distribution, is less likely to be detected but requires a higher proportion of poisoned data to reach similar levels of attack efficacy.
This highlights a crucial adversarial trade-off: maximizing attack effectiveness with high-contrast triggers or prioritizing stealth and applicability with subtler triggers, albeit at the cost of efficiency.

We also provide evidence that entropy-based backdoor attacks are considerably more effective than random attacks at test time, leading to a more pronounced decrease in policy performance at test time. 
This suggests that selectively triggering the backdoor in states of high uncertainty is a powerful strategy, particularly in scenarios where the attacker cannot trigger the backdoor frequently. Future work should explore whether there exist optimal trigger timing attacks for imitation learning. We also hypothesize that if the adversary could use access to a digital twin of the environment to better optimize the timing of backdoor trigger injections.

In conclusion, we demonstrate that behavioral cloning (BC) policies are highly vulnerable to minimal clean-label backdoor dataset poisoning attacks. As the first to investigate this vulnerability, our work highlights the need for further research into the security challenges of BC. In particular, developing backdoor detection and removal techniques to enhance the robustness and security of imitation learning methods, is an especially exciting area for future work.


    


\bibliographystyle{splncs04}
\bibliography{bc_attacks}

\newpage
\appendix
\section{Appendix}
\subsection{Additional Patch Size Ablation}
\label{app:more-patch-size}







Our results in Subsection~\ref{sec:patch-size-main} in the main paper show that small to moderate patch sizes (e.g., 5x5 to 25x25) occasionally yield higher mean rewards than the unpatched baseline. We hypothesize that the backdoor trigger acts as a decisive cue, particularly for the gas action. When BC policies hesitate or waver due to compounding errors, a strong and consistent visual pattern reduces ambiguity and stabilizes the gas prediction, which improves the forward progress and reward.

Replacing the red square with a Gaussian patch at the same poisoning rate yields a qualitatively similar efficacy curve where the control increases slightly with $N$ and plateaus at a small $N$. However, the absolute control rate is lower and more variable than with the red patch, as a Gaussian trigger is harder to learn at only 5\% poisoning. The reward trends remain within a narrow band of the unpatched baseline.

Beyond this local boost, enlarging the patch past a small threshold yields no further gains in control accuracy. The trigger reaches near perfect reliability with a 5$\times$5 or 10$\times$10 patch. Larger patches do not improve performance and may increase detectability. This indicates a minimal effective patch size for attackers, the point where the trigger is already linearly separable for early convolutional layers, after which bigger patches have negligible impact on learnability.

Existing literature~\cite{gu2019badnets} reveals trade-offs between the two objectives of attack success and stealth, where injecting a larger patch to induce stronger spurious correlations is more likely to be detected or occludes task‑relevant pixels. Conversely, smaller triggers can remain visually insignificant to humans yet may require a higher poisoning rate for success.  

To study this balance, we sweep the square patch edge length $N\in\{1:96\}$ pixels while holding the poisoning percentage fixed at \textit{5\%}, as we saw previously that this low poisoning rate can yield near perfect adversarial results.  
For each $N$ we keep the trigger color red, location on the top left of the frame, and target action for gas identical, training 5 independent BC policies per condition.
\begin{figure}[t]
\captionsetup{belowskip=-5pt}
    \centering
    \includegraphics[width=\textwidth]{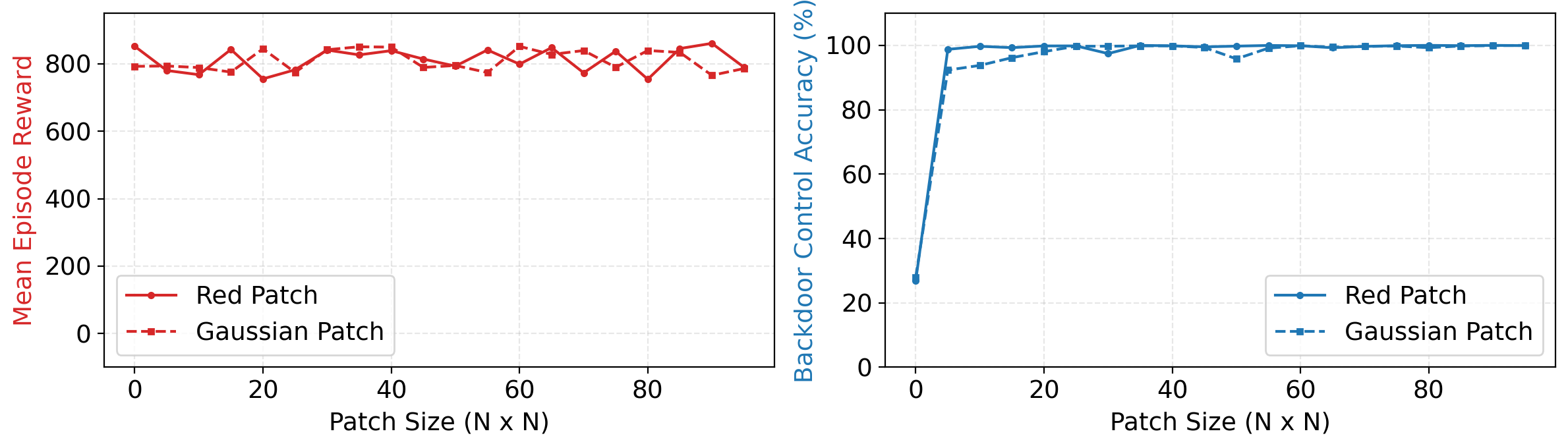}
    \caption{\textbf{Patch-size sweep comparison at a higher poisoning rate (20\% of gas-labeled frames) for red and Gaussian patches}. On the left, we see the mean reward being comparable between both patch types, while on the right we see a high control rate via the patch across the board. The red patch outperforms the Gaussian patch at smaller patch sizes, but the Gaussian patch still performs well with over 90\% attack accuracy.}
    \label{fig:patch-size-20}
\end{figure}

We repeat the sweep at a higher poisoning rate of 20\% of gas-labeled frames to rule out a low-rate artifact. For both red and Gaussian triggers, control saturates with very small patches $N$=5-10 and remains flat afterward. The mean reward stays within a narrow band around the baseline across all $N$. These results confirm a minimal effective patch size exists where choosing a smaller $N$ can reduce detectability and occlusion without reducing attack efficacy.


\subsection{Type of Patch}
\label{app:more-patch-type}
\begin{figure}[t]
    \centering
    \begin{subfigure}[t]{0.24\textwidth}
        \centering
        \includegraphics[width=\linewidth]{Springer_Latex_Template/imgs/poisoned_clean.png}
        \caption{\scriptsize Clean}
    \end{subfigure}%
    \hfill
    \begin{subfigure}[t]{0.24\textwidth}
        \centering
        \includegraphics[width=\linewidth]{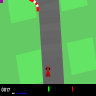}
        \caption{\scriptsize Color Space}
    \end{subfigure}%
    \hfill
    \begin{subfigure}[t]{0.24\textwidth}
        \centering
        \includegraphics[width=\linewidth]{Springer_Latex_Template/imgs/poisoned_gauss.png}
        \caption{\scriptsize Gaussian}
    \end{subfigure}%
    \hfill
    \begin{subfigure}[t]{0.24\textwidth}
        \centering
        \includegraphics[width=\linewidth]{Springer_Latex_Template/imgs/poisoned_red.png}
        \caption{\scriptsize Red}
    \end{subfigure}
    
    \caption{Visual comparison of poisoned frames using different attack types. All attacks shown are a 3$\times$3 patch in the top left corner of the visual observation.}
    \label{fig:poisoned-comparison}
\end{figure}

Finally, we compare different synthetic patch triggers to understand which visual characteristics most influence the effectiveness of the backdoor. Specifically, we consider three designs: a \textbf{fixed-color square}, a \textbf{Gaussian noise patch}, and a \textbf{color-space shift patch}. The different patches are visualized within poisoned frames in~\autoref{fig:poisoned-comparison}.

The \textbf{fixed-color square} is a red patch which is high contrast and spatially consistent, which leads to early convolutional layers learning it easily. Its main drawback is its perceptibility, since the uniform bright color is easily detected by human reviewers and automated color-based defenses.


The \textbf{Gaussian noise patch} is sampled from $\mathcal{N}(127, 30^2)$, clipped to $[0, 255]$, and kept fixed for the entire training process. We generate it once and reuse the same instance on every poisoned frame to provide a consistent correlation (resampling for each frame would make the trigger stochastic and difficult to memorize). Its main benefit is a lower chance of detection by simple defenses, while its drawback is that it is harder to learn at lower poisoning rates and may require a higher rate to be more effective.


The \textbf{color-space shift patch} is a trigger where we increase all RGB channel values in that region by a fixed offset of +40, then clip to valid RGB values. This distorts color balance and does not remain static through the rollout. It is stealthy and can resemble natural lighting or tint changes, while its drawback is reduced learnability, especially when the patch overlaps low-contrast regions.


\begin{table}[t]
\centering
\small
\caption{Comparison of patch types at 5\% poisoned gas frames. Backdoor control accuracy is the percentage of poisoned observations predicted as ``gas''.}
\label{table:app-patch-type}
    \begin{tabular}{lcccccc}
    \toprule
    Patch Type &&& Mean Reward &&& Backdoor Control Acc. (\%) \\
    \midrule
    No patch             & &  & 860.85 ± 86.30 &&& 22.33 \\
    Red Patch            &  &     & 832.66 ± 94.99 &&& 97.86 \\
    Gaussian Noise Patch &&& 847.79 ± 74.39 &&& 85.07 \\
    Color-Space Shift    &&& 878.41 ± 60.10 &&& 88.13 \\
    \bottomrule
    \end{tabular}
\end{table}

Our results at 5\% poisoning are shown in~\autoref{table:patch-type}. All three patch types achieve high backdoor control at 5\% of poisoned gas-labeled frames. The red patch attains the highest control rate because its sharp color and structure are easily learned by convolutional filters. It also yields the highest mean reward, suggesting more stable throttle behavior during rollouts.


The Gaussian noise patch and the color-space shift achieve slightly lower control rates, though still high enough to compromise trained models. Their lower contrast and subtle texture make them harder for the model to isolate at low poisoning budgets, and in further experiments they match the red patch only when the poisoning rate is higher. Their mean rewards are slightly lower than the red patch but remain close to the clean baseline, indicating that subtle triggers are learnable by behavioral cloning models with somewhat reduced robustness and greater variability in recognition.

\begin{figure}[t]
\captionsetup{belowskip=-15pt}
    \centering
    \includegraphics[width=\textwidth]{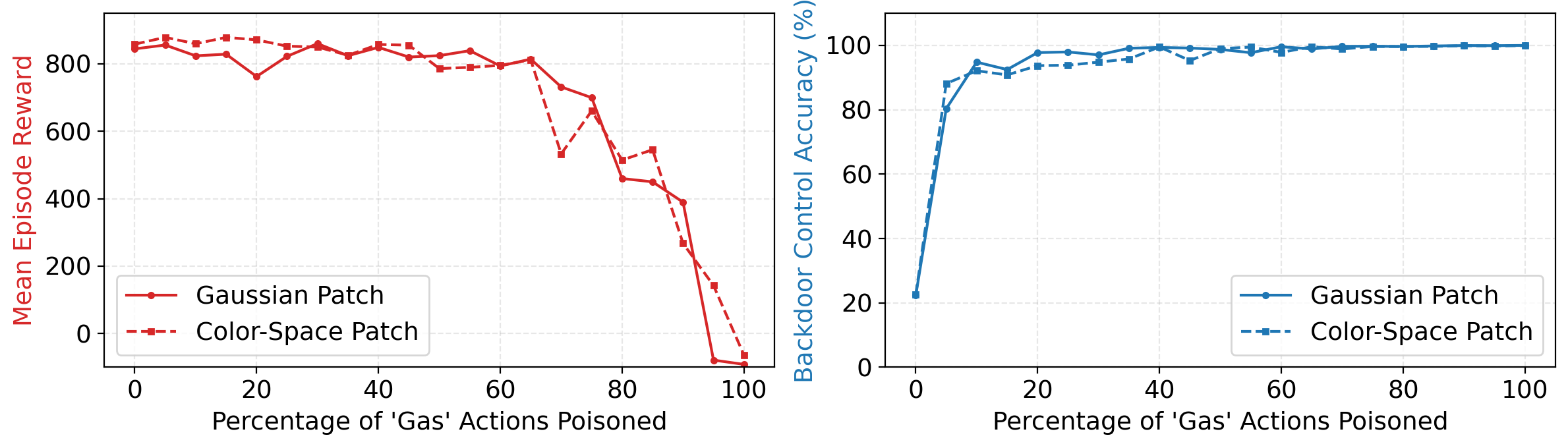}
    \caption{\textbf{Poisoning‑rate sweeps for Gaussian and color-space triggers.} Both patch types show similar performance across different 'gas' action poison percentages. We see that the accuracy of the backdoor patch seems to be higher for the gaussian patch at lower percentages, while the mean reward and backdoor accuracy for the rest of the poison percentages for both patches follow the same trend.}
    \label{fig:triggertype-rate-sweeps}
\end{figure}

To examine how the subtler triggers scale, we sweep the fraction of gas-labeled frames poisoned with the Gaussian and color-space triggers. As shown in~\autoref{fig:triggertype-rate-sweeps}, both triggers improve with higher poisoning rates. The Gaussian patch reaches near perfect control at $20$–$25\%$ while the color-space patch reaches similar levels of control around $20$-$25\%$. Very high poisoning rates reduce the mean reward, consistent with over-biasing toward the gas action. 


We conclude that the trade-off between stealth (Gaussian and color-space patches vs. a solid red patch) and performance exists, but is not as significant as expected. Future works may focus on creating a realistic backdoor trigger patch that is more subtle to the human eye, even though it may require a higher poisoning rate to be effective. 

\end{document}